\documentclass[runningheads]{llncs}
\usepackage{graphicx}
\usepackage{tikz}
\usepackage{comment}
\usepackage{amsmath,amssymb} 
\usepackage{bm}
\usepackage{anyfontsize}
\usepackage{xcolor}
\usepackage{booktabs}

\usepackage[ruled, lined,noend, linesnumbered, commentsnumbered]{algorithm2e}

\DeclareMathOperator*{\argmin}{arg\,min}

\definecolor{commentsColor}{RGB}{219, 48, 122}
\newcommand\mycommfont[1]{\footnotesize\ttfamily\textcolor{commentsColor}{#1}}
\SetCommentSty{mycommfont}
\let\oldnl\nl
\newcommand{\nonl}{\renewcommand{\nl}{\let\nl\oldnl}}

\usepackage[normalem]{ulem}
\useunder{\uline}{\ul}{}

\graphicspath{{Figs/}}

\usepackage[accsupp]{axessibility}  

\newcommand{\etal}{\emph{et al.}}
\newcommand{\ie}{\emph{i.e.}}
\usepackage[pagebackref,breaklinks,colorlinks]{hyperref}

\usepackage[capitalize]{cleveref}
\crefname{section}{Sec.}{Secs.}
\Crefname{section}{Section}{Sections}
\Crefname{table}{Table}{Tables}
\crefname{table}{Tab.}{Tabs.}

\usepackage[format=plain,labelformat=simple,labelsep=period,font=small,compatibility=false]{caption}
\usepackage[font=scriptsize,skip=3pt,subrefformat=parens]{subcaption}

\usepackage[symbol]{footmisc}

\definecolor{myyellow}{rgb}{1, 0.812, 0}
\definecolor{mygreen}{rgb}{0.0667, 0.729, 0.271}
\definecolor{myred}{rgb}{0.729, 0.0902, 0.0667}

\begin{document}
\pagestyle{headings}
\mainmatter
\def\ECCVSubNumber{5080}  

\title{PrivHAR: Recognizing Human Actions From Privacy-preserving Lens} 

\titlerunning{PrivHAR}
%
\author{Carlos Hinojosa\inst{1,2,*} \and
Miguel Marquez\inst{1}\and
Henry Arguello\inst{1}\and
Ehsan Adeli\inst{2}\and
\\ Li Fei-Fei\inst{2}\and
Juan Carlos Niebles\inst{2}}
\authorrunning{C. Hinojosa et al.}
%
\institute{Universidad Industrial de Santander, Colombia\\ \and
Stanford University, USA\\
{\tt\small \url{https://carloshinojosa.me/project/privhar/}}
\vspace{-10pt}
}

\maketitle
\footnotetext[1]{\email{carlos.hinojosa@saber.uis.edu.co}}
\begin{abstract}
The accelerated use of digital cameras prompts an increasing concern about privacy and security, particularly in applications such as action recognition.   
In this paper, we propose an optimizing framework to provide robust visual privacy protection along the human action recognition pipeline. 
Our framework parameterizes the camera lens  to successfully degrade the quality of the videos to inhibit privacy attributes and protect against adversarial attacks while maintaining relevant features for activity recognition.
We validate our approach with extensive simulations and hardware experiments. 
\end{abstract}

\section{Introduction}
\label{sec:intro}

\begin{figure}[t]
	\centering
	\includegraphics[width=0.8\columnwidth]{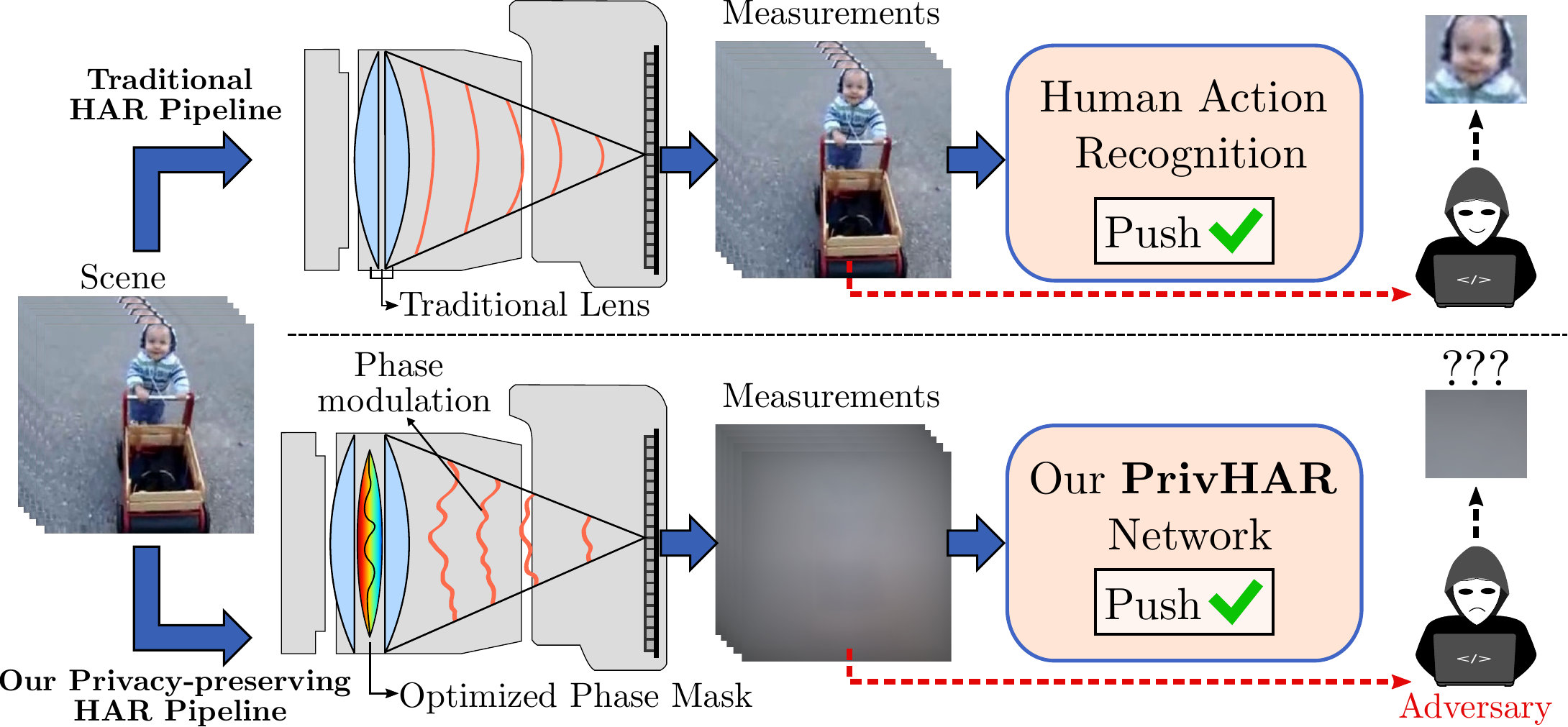} \vspace{-5pt}
	\caption{\small Traditional HAR pipeline uses standard cameras that acquire visual details from the scene leading to privacy issues. We introduce PrivHAR, an adversarial optimization framework that learns a lens' phase mask to encode human action features and perform HAR while obscuring privacy-related attributes.}\vspace{-15pt}
	\label{fig:intro}
\end{figure}



We are at the beginning of a new era of smart systems. From health care to video games, computer vision applications have provided successful solutions to real-world problems \cite{bommasani2021opportunities,krishna2021visual,liu2020spatiotemporal}. For decades, cameras have been engineered to imitate the human vision system, and machine learning algorithms are always constrained to be optimized using high-quality images as inputs. However, the abundance and growing uses of smart devices are also causing a social dilemma: we want intelligent systems (e.g., in our home) to recognize relevant events and assist us in our activities, but we also want to ensure they protect our privacy.

There have been some previous studies dealing with such a social dilemma. For instance, some early works rely on hand-crafted strategies, e.g., pixelation \cite{van2009dimensionality}, blurring \cite{padilla2015visual}, face/object replacement \cite{chen2007tools}, and person de-identification \cite{agrawal2011person}, to degrade sensitive information. More recently, Ren \etal~\cite{ren2018learning} proposed an adversarial training strategy to learn to anonymize faces in videos and then perform activity detection. Similarly, using adversarial training, \cite{wu2020privacy,wu2018towards} proposed to optimize privacy attributes and recognition performance. However, all these methods rely on software-level processing of original high-resolution videos, which may already contain privacy-sensitive data. Hence, there is a possibility of these original videos being snatched by an attacker. Instead of developing new algorithms or designing software-level solutions that still rely on high-resolution images and videos as input, we believe that the privacy-preserving problem in computer vision should be addressed directly within the camera hardware, \ie, sensible visual data should be protected before the images are acquired in the sensor.

Currently, few works have been developed in this direction. For instance, \cite{ryoo2018extreme,ryoo2017privacy} proposed to use low-resolution cameras to create privacy-preserving anonymized videos and perform human action recognition. Also, Pittaluga \etal~\cite{pittaluga2015privacy} introduced a defocusing lens to provide a certain level of privacy over a working region. On the other hand, several works used depth cameras to protect privacy and perform human action recognition \cite{ji2018skeleton,ahmad2019human}. These approaches rely on a fixed optical system; thus, their main contribution included designing an algorithm for a specific input type. More recently, \cite{Hinojosa_2021_ICCV} proposed to jointly design the lens of a camera and optimize a deep neural network to achieve two goals: privacy protection and human pose estimation. However, the formulation of the optimization in this work poses different problems: the privacy-preserving loss is not bounded as it maximizes an $\ell_2$ term to enforce degradation, which may cause instability in the optimization; authors only used one human pose estimation model (OpenPose \cite{cao2019openpose}) in all experiments and its not clear if the method works with other pose estimators. More importantly, to test and measure privacy, authors performed adversarial attacks after training the network; hence such attacks were not considered in the lens design.

In this paper, we address the problem of privacy-preserving human action recognition and propose a novel adversarial framework to provide robust privacy protection along the computer vision pipeline, see Fig. \ref{fig:intro}. We adopt the idea of end-to-end optimization of the camera lens and vision task \cite{Hinojosa_2021_ICCV,metzler2020deep,sitzmann2018end} and propose an optimization scheme that: \textbf{(1)} Incorporates adversarial defense objectives into the learning process across a diversity of canonical privacy categories, including face, skin color, gender, relationship, and nudity detection. \textbf{(2)} Encourages distortions in the videos without compromising the training stability by including the structural similarity index (SSIM) \cite{hore2010image} in our optimization loss. \textbf{(3)} To further preserve the temporal information in the distorted videos, we use temporal similarity matrices (TSM) and constrain the structure of the temporal embeddings from the private videos to match the TSM of the original video.


We test our approach with two popular human action recognition backbone networks. To experimentally test our privacy-preserving human action recognition network (PrivHAR) and lens design, we built a proof-of-concept optical system in our lab. Our testbed  acquires distorted videos and their non-distorted version simultaneously. Our experimental results in hardware match the simulations. While we do observe a trade-off between Human Action Recognition (HAR) accuracy and image distortion level, our proposed PrivHAR system  offers robust protection with reasonable accuracy.

\vspace{-14pt}
\section{Related Work}
\vspace{-10pt}
Human action recognition is a challenging task \cite{pareek2021survey} and has many applications, such as video surveillance, human-computer interfaces, virtual reality, video games, and sports action analysis. Therefore, developing privacy-preserving approaches for HAR is even more challenging and has not been widely explored.


\noindent\textbf{Human Action Recognition (HAR).}
Nowadays, there are multiple approaches in the computer vision literature for addressing the HAR problem. Some prior work relies on 2D CNNs to conduct video recognition \cite{chollet2017xception,karpathy2014large,wang2016temporal,christoph2016spatiotemporal}. 
A major drawback of 2D CNN approaches is not properly modeling the temporal dynamics. On the other hand, 3D CNN-based approaches use spatial and temporal convolutions over the 3D space to infer complicated spatio-temporal relationships. For instance, C3D~\cite{tran2015learning} is a 3D CNN based on the VGG model that learns spatio-temporal features from a frame sequence. However, 3D CNNs are typically computationally heavy, making the deployment difficult. Therefore, many efforts on HAR focus on proposing new efficient architectures; for example, by decomposing 3D filters into separate 2D spatial and 1D temporal filters \cite{kopuklu2019resource,tran2019video} or extending efficient 2D architectures to 3D counterparts. Moreover, RubiksNet~\cite{fan2020rubiksnet} is a hardware-efficient architecture for HAR based on a shift layer that learns to perform shift operations jointly in spatial and temporal context. We build our proposed PrivHAR using both C3D and RubiksNet.




\noindent\textbf{Privacy protection in Computer Vision.} Currently, few works address the privacy-preserving HAR problem. We divide prior work into software-level and hardware-level protection, where we consider the latter more robust to attacks.

\textit{Software-level Privacy-preserving HAR.} Most prior privacy-preserving works apply different computer vision algorithms to the video data after their acquisition. The literature has relied on domain knowledge and hand-crafted approaches, such as pixelation, blurring, and face/object replacement, to protect sensitive information \cite{agrawal2011person,chen2007tools,padilla2015visual}. These methods can be useful in settings when we know in advance what to protect in the scene. More recent works propose a more general approach that learns privacy-preserving encodings through adversarial training \cite{wu2020privacy,brkic2017know,pittaluga2019learning}. These methods learn to degrade or inhibit privacy attributes while maintaining important features to perform inference tasks and provide more robust videos to adversarial attacks. 
Ren \etal~\cite{ren2018learning} use adversarial training to learn a video anonymizer and remove facial features for activity detection. Similarly, Wu \etal~\cite{wu2020privacy,wu2018towards} proposed an adversarial framework that learns a degradation transform for the video inputs using a 2D convolution layer. These works optimize the trade-off between action recognition performance and the associated privacy budget on the degraded video. Although these software-level approaches preserve privacy, the acquired images are not protected.


\textit{Hardware-level Privacy-preserving HAR.} These approaches rely on the camera hardware to add an extra layer of security by removing sensitive data during the imaging sensing. Prior hardware-level privacy-preserving approaches use low-resolution cameras to anonymize videos, i.e., the videos are intentionally captured to be in special low-quality conditions that only allow for the recognition of some events or activities while avoiding the unwanted leak of the identity information for the human subjects in the video \cite{purwanto2019extreme,ryoo2017privacy}. In \cite{pittaluga2016pre,pittaluga2015privacy}, two optical designs were proposed that filter or block sensitive information directly from the incident light-field before sensor measurements acquisition, enabling k-anonymity and privacy protection by using a camera with defocusing lens. In particular, they show how to select a defocus blur that provides a certain level of privacy over a working region within the sensor size limits; however, only using optical defocus for privacy may be susceptible to reverse engineering. In addition, the authors did not test their method on the action recognition task. More recently, \cite{wang2019privacy} proposed a coded aperture camera system to perform privacy-preserving HAR directly from encoded measurements without the need for image restoration. However, it was only tested for indoor settings in a small dataset.



\begin{figure*}[t]
	\centering
	\includegraphics[width=\linewidth]{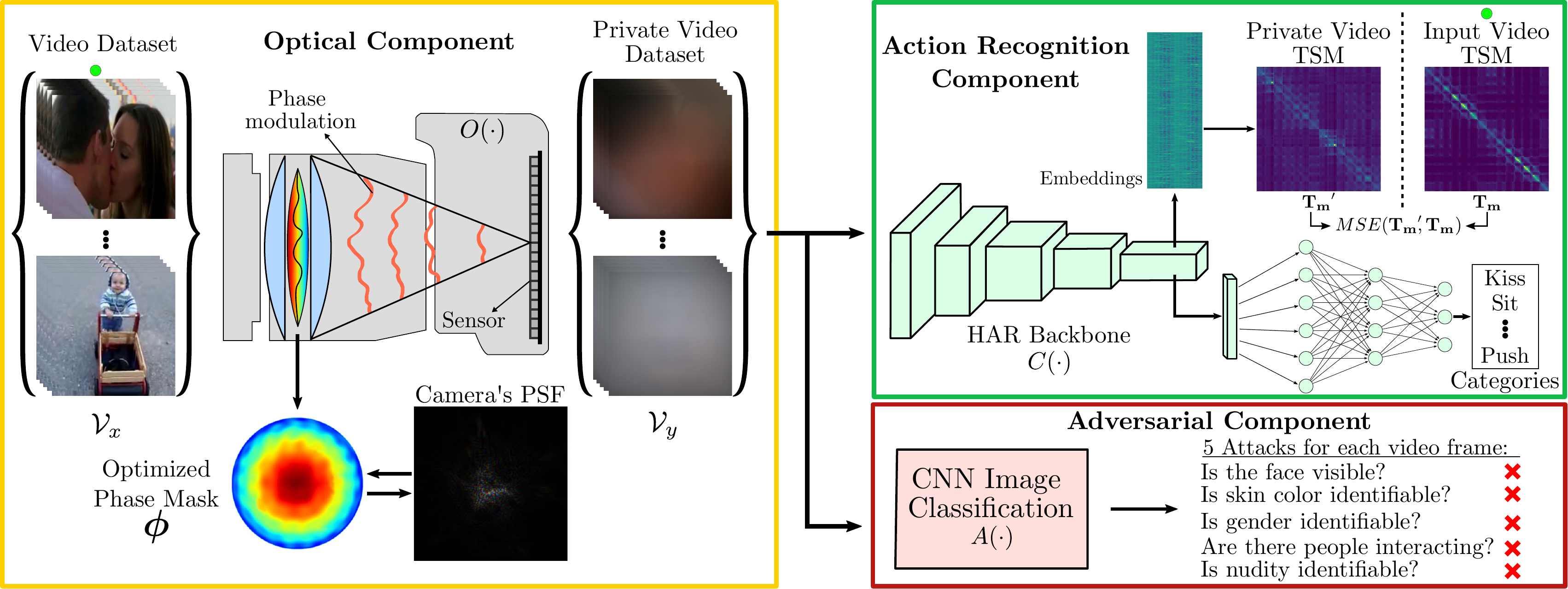} \vspace{-15pt}
	\caption{\small Our proposed end-to-end framework. Our \textcolor{myyellow}{optical component} consists of a camera with two thin convex lenses and a phase mask between them. We achieve robust privacy protection by training an adversarial framework under three goals: (1) to learn to add aberrations to the lens surface such that the acquired videos are distorted to obscure private attributes while still preserving features to (2) achieve high \textcolor{mygreen}{video action recognition} accuracy, and (3) being robust to \textcolor{myred}{adversarial attacks}.}\vspace{-14pt}
	\label{fig:proposedM}
\end{figure*}

\vspace{-10pt}
\section{Privacy-preserving Action Recognition}
\vspace{-5pt}

We are interested in human action recognition from privacy-preserving videos. We propose a framework to accomplish three goals:  1) to learn the parameters of a robust privacy-preserving lens by backpropagating the gradients from the action recognition and adversarial branches to the camera; 2) to learn the parameters of an action recognition network to perform HAR on the private videos with high accuracy; 3) to obtain private videos that are robust to adversarial attacks. 
Our framework (\cref{fig:proposedM}) consists of three parts: optical, action recognition, and an adversarial component.

The optical component consists of a camera with two thin convex lenses and a phase mask between them. Our simulated camera takes a video $\mathbf{V_x}\in \mathbb{R}_{+}^{w\times h \times3 \times T}=\left\{\mathbf{X}_t\right\}_{t=1}^{T}$ as input, which has $w\times h$ pixels and $T$ frames, and outputs the corresponding distorted video $\mathbf{V_y}\in \mathbb{R}_{+}^{w\times h \times3 \times T}$. Formally, $\mathbf{V_y}=O(\mathbf{V_x})$, where we denote our designed camera as the function $O(\cdot)$, which distorts every single frame $\mathbf{X}_t \in \mathbb{R}_{+}^{w\times h \times3}$, and produces the respective \textit{private} frames $\mathbf{Y}_t \in \mathbb{R}_{+}^{w\times h \times3}$. Then, the distorted video $\mathbf{V_y}$ passes through the action recognition component where a convolutional neural network $C$ predicts the class labels. Besides, $\mathbf{V_y}$ also passes through the adversarial component where an attribute estimator network $A$ tries to predict the private information (attributes) from the distorted video. All three components consist of neural networks with trainable parameters, and the whole framework is trained adversarially. At the end of the optimization process, we obtain the optimal camera lens parameters $\theta_{o}^{*}$, and the optimal action recognition parameters $\theta_{c}^{*}$. Hence, the loss function of our adversarial framework is formulated as follows:
{\small
 	\setlength{\belowdisplayskip}{3pt}
 	\setlength{\abovedisplayskip}{3pt}
	\begin{equation}
		\theta_{o}^{*}, \theta_{c}^{*} =  \argmin_{\theta_{o}, \theta_{c}} L(O) + L(C) - L(A),
		\label{eq:general_loss}
	\end{equation}
}where $L(O)$, $L(C)$, and $L(A)$ are the loss functions for our optical component, action recognition component, and adversarial component, respectively.

During inference, we can construct a camera lens using the optimal parameters $\theta_{o}^{*}$ that acquires degraded images, on which our network $C$ can perform HAR. Since we develop our protection directly in the optics (camera lens), it provides an extra layer of protection and, hence, is more difficult for a hacker to attack our system to reveal the person's identity. One could also deploy a less secure software-only approach implementing image degradation post-acquisition. A hybrid solution consists of designing an embedded chipset responsible for distorting the videos immediately after the camera sensor. 
\vspace{-15pt}
\subsection{Optical Component}
\label{subsec:hardware_layer}
\vspace{-5pt}

The main goal of the optical component in our adversarial framework PrivHAR (Fig. \ref{fig:proposedM}) is to design a phase mask to visually distort videos (hence obscuring privacy-sensitive attributes), encode the physical characteristics and preserve human action features to perform HAR. We adopted a similar strategy as the authors in \cite{sitzmann2018end,Hinojosa_2021_ICCV} to couple the modeling and design of two essential operators in the imaging system: wave propagation and phase modulation.

\noindent\textbf{Image Formation Model.} We model the image acquisition process using the point spread function (PSF) defined in terms of the lens surface profile to emulate the wavefront propagation and train the parameters of the refractive lens. Considering by the Fresnel approximation and the paraxial regime \cite{goodman2005introduction}, for incoherent illumination, the PSF can be described by 
{\small
	\setlength{\belowdisplayskip}{3pt}
	\setlength{\abovedisplayskip}{3pt}
	\begin{eqnarray}
		H(u',v') = |\mathcal{F}^{-1}\{\mathcal{F}\{t_L(u,v)\cdot t_\phi(u,v)\cdot W(u,v)\}\cdot T(f_u,f_v)\}|^2,
		\label{eq:psf_continuous}
	\end{eqnarray}
}where $W(u,v)$ is the incoming wavefront, $T(\cdot)$ represents the transfer function with $(f_u,f_v)$ as the spatial frequencies, $t_\phi(u,v)=\exp(-ik\phi(u,v))$ with $\phi(u,v)$ as the lens phase mask and $k=2\pi/\lambda$ as the wavenumber, $t_L(\cdot)$ denotes the light wave propagation phase with $t_L(u,v)=\exp \left( -i\frac{k}{2z}(u^2+v^2) \right)$ with $z$ as the object-lens distance, $\mathcal{F}\{\cdot\}$ denotes the 2D Fourier transform, and $(u',v')$ is the spatial coordinate on the camera plane. The values of $\phi(\cdot)$ are modelled via the Zernike polynomials with $\phi(u,v)=\mathcal{R}_{\bar{n}}^{\bar{m}}(\sqrt{u^2+v^2})\cdot\cos{(\arctan{(v/u)})}$, where $\mathcal{R}\left(\cdot\right)$ represents the radial polynomial function \cite{lakshminarayanan2011zernike}, $\bar{m}$ and $\bar{n}$ are nonnegative integers with $\bar{n} \geq \bar{m} \geq 0$. To train the phase mask values using our PrivHAR, we discretize the phase mask $\phi(\cdot)$ as:
{\small 
	\setlength{\belowdisplayskip}{1pt}
	\setlength{\abovedisplayskip}{1pt}
	\begin{equation}
		\boldsymbol{\phi}=\sum_{j=1}^{q} \alpha_j \mathbf{Z}_j,
		\label{eq:zernike_model}
	\end{equation}
}where $\mathbf{Z}_j$ denotes the $j$-th Zernike polynomial in Noll notation, and $\alpha_j$ is the corresponding coefficient \cite{born2013principles}. Each Zernike polynomial describes a wavefront aberration \cite{lakshminarayanan2011zernike}; hence the phase mask $\boldsymbol{\phi}$ is formed by the linear combination of all aberrations. In this regard, the optical element parameterized by $\boldsymbol{\phi}$ can be seen as an optical encoder, where the coefficients $\alpha_j$ determine the data transformation. Therefore, our adversarial training finds a set of coefficients $\theta_{o}^* = \left\{\alpha_j\right\}_{j=1}^{q}$ that provides the maximum visual distortion of the scene but allows to extract relevant features to perform HAR. Using the defined PSF-based propagation model (assuming that image formation is a shift-invariant convolution of the image and PSF), the acquired private images for each RGB channel can be modelled as:
{\small
	\setlength{\belowdisplayskip}{5pt}
	\setlength{\abovedisplayskip}{5pt}
	\begin{equation}
		\mathbf{Y}_{\ell} =\mathcal{G}_{\ell}\left(\mathbf{H}_{\ell}*\mathbf{X}_{\ell}\right) + \boldsymbol{\eta}_{\ell},
	\end{equation}
}where $\mathbf{X}_{\ell} \in \mathbb{R}_{+}^{w\times h}$ represents the discrete image from the $\ell$ channel, with each pixel value in $\left[0,1\right]$; $\mathbf{H}_{\ell}$ denotes the discretized version of the PSF \cite{goodman2005introduction} in Eq. (\ref{eq:psf_continuous}) for the channel $\ell$, $\boldsymbol{\eta}_{\ell}\in\mathbb{R}^{w\times h}$ represents the Gaussian noise in the sensor, and $\mathcal{G}_{\ell}(\cdot):\mathbb{R}^{w\times h}\rightarrow \mathbb{R}^{w\times h}$ is the camera response function, which is modeled as a linear function. Please see our supplementary document for a schematic diagram of the light propagation in our model.


\textbf{Loss Function.}
To encourage image degradation, we train our network to minimize the quality of the acquired image by our camera $\mathbf{Y}=\left\{\mathbf{Y}_\ell\right\}_{\ell=1}^{3}$ in comparison with the original image $\mathbf{X}=\left\{\mathbf{X}_\ell\right\}_{\ell=1}^{3}$. Instead of maximizing the $\ell_2$ norm error between the two images as previous works did \cite{Hinojosa_2021_ICCV}, we use the structural similarity index (SSIM) \cite{wang2004image} in our optimization loss to measure quality. The $\ell_2$ norm does not have an upper bound; hence maximizing it to enforce degradation causes instability in the optimization. On the other hand, the SSIM function is bounded, which leads to better stability during training. Specifically, the SSIM value ranges between 0 and 1, where values near 1 (better quality) indicate more perceptual similarity between the two compared images. Then, we define the loss function for our camera lens optimization as:
{\small
	\setlength{\belowdisplayskip}{3pt}
	\setlength{\abovedisplayskip}{3pt}
	\begin{equation}
		L(O) \triangleq SSIM(\mathbf{X}, \mathbf{Y}).
	\end{equation}}
Since we encourage distortion in the camera's output images/videos, the $L(O)$ loss is minimized in our adversarial training algorithm, see \cref{alg:adversarial_training}.

\vspace{-10pt}
\subsection{Action Recognition Component}
\vspace{-2pt}


We can use any neural network architecture in our adversarial framework to perform human action recognition. In this work, without loss of generality, we adopt two HAR CNN architectures: the well-known C3D \cite{tran2015learning}, and the Rubkisnet \cite{fan2020rubiksnet}, a more recent and efficient architecture for HAR. For a set of private videos, we assume that the output of the classifier $C$ is a set of action class labels $\mathcal{S}_C$. Then, we can use the standard cross-entropy function $\mathcal{H}$ as the classifier's loss.

On the other hand, since our degradation model distorts each frame of the input video separately (2D convolution), part of the temporal information could be lost, decreasing the performance of the HAR CNN significantly. To preserve temporal information, we use temporal similarity matrices (TSMs). TSMs are useful representations for human action recognition and have been employed in several works \cite{junejo2010view,dwibedi2020counting,panagiotakis2018unsupervised,sun2015exploring} due to their robustness against dynamic view changes of the camera when paired with appropriate feature representation. Unlike previous works, we propose using TSMs as a proxy to keep the temporal information (features) similar after distortion: we build a TSM for the original and private videos and compare their structures. Specifically, we take the embeddings $\mathbf{\hat{e}}$ from the last convolutional layer of our HAR CNN architecture and compute the TSM values using the negative of the squared euclidean distance, i.e., $(\mathbf{T_m}')_{n_1n_2}=-\|\mathbf{\hat{e}}_{n_1}-\mathbf{\hat{e}}_{n_2}\|^2$. Then, we calculate the mean square error (MSE) between the $\mathbf{T_m}'$ and the TSM from the input video $\mathbf{T_m}$, which was computed similarly using the last convolutional layer of the corresponding pretrained HAR CNN (non-privacy) network. We define the action recognition objective as:
{\small
	\setlength{\belowdisplayskip}{5pt}
	\setlength{\abovedisplayskip}{5pt}
	\begin{equation}
		L(C) \triangleq \mathcal{H}(\mathcal{S}_C,C(\mathcal{V}_y)) + MSE(\mathbf{T_m},\mathbf{T_m}'),
	\end{equation}
}where $\mathcal{V}_y$ denotes the set of $E$ private videos:
$\mathcal{V}_y=\left\{\mathbf{V_y}^{e}\right\}_{e=1}^{E}=\left\{O(\mathbf{V_x}^{e})\right\}_{e=1}^{E}$.

\vspace{-10pt}
\subsection{Adversarial Component and Training Algorithm}
\vspace{-2pt}

The attacks that an adversarial agent could perform to our privacy-preserving pipeline depends on the definition of privacy. There are different ways to measure privacy and this is, in general, not a straightforward task. For example, in smart homes with video surveillance, one might often want to avoid disclosure of the face or identity of persons. Therefore, an adversarial agent could try to attack our system by training a face detection network. However, there are other privacy-related attributes, such as race, gender, or age, that an adversarial agent could also wanted to attack too. In this work, we define the adversarial attack as a classification problem, where a CNN network $A$ takes a private video $\mathbf{V_y}$ as input and tries to predict the corresponding private information. Therefore, the goal of our adversarial training is to try that the predictions from $A$ diverges from the set of class labels $\mathcal{S}_A$ that describe the private information within the scene. To train the attribute estimator network, we also use the cross-entropy $\mathcal{H}$ function and define the adversarial loss as:
{\small
	\setlength{\belowdisplayskip}{5pt}
	\setlength{\abovedisplayskip}{5pt}
	\begin{equation}
		L(A) \triangleq \mathcal{H}(\mathcal{S}_A,A(\mathcal{V}_y)).
\end{equation}}


\Cref{alg:adversarial_training} summarizes the proposed adversarial training scheme. \underline{Before} performing the adversarial training, we first train each framework component separately without privacy concern to obtain the optimal performance on each task. Specifically, we train the optical component $O$ by minimizing $1-L(O)$ to acquire videos without distortions, i.e., $\mathbf{V_y}$ videos are very similar to the corresponding input $\mathbf{V_x}$. We also train the HAR network $C$ by minimizing $\mathcal{H}(\mathcal{S}_C, C(\mathcal{V}_x))$, obtaining the highest action recognition accuracy (the upper bound). Finally, we train the attribute estimator network $A$ by minimizing $\mathcal{H}(\mathcal{S}_A, A(\mathcal{V}_x))$, thus obtaining the highest classification accuracy (the upper bound). After initialization, we start the adversarial training shown in \cref{alg:adversarial_training},  where, for each epoch and every batch, we first acquire the private videos with our camera $O$. Then, we update the parameters of the camera $\theta_{o}$ by freezing the attribute estimator network parameters $\theta_{a}$ and minimizing the weighted sum $L(O) + \gamma_1L(C) - \gamma_2L(A)$, shown on line 4 of the algorithm. Similarly, we update the parameters of the HAR network $\theta_{c}$ by freezing the attribute estimator network parameters and using the private videos acquired on line 3 to minimize $L(C)$. Finally, we perform the adversarial attack by minimizing $L(A)$ and updating the parameters of the attribute estimator network $\theta_{a}$ while the camera and HAR network parameters are fixed. Contrary to the prior work \cite{Hinojosa_2021_ICCV}, our training scheme jointly models the privacy-preserving optics with HAR and adversarial attacks during training.

\setlength{\textfloatsep}{0.1cm}
\begin{algorithm}[t]
	\SetKwFunction{isOddNumber}{isOddNumber}
	\footnotesize
	\SetKwInOut{KwIn}{Input}
	\SetKwInOut{KwOut}{Output}
	\KwIn{Video Dataset $\mathcal{V}_x=\left\{\mathbf{V_x}^{e}\right\}_{e=1}^{E}$. Hyperparameters $\beta_{o}, \beta_{c}, \beta_{a}, \gamma_1, \gamma_2$}
	\KwOut{$\theta_{o},\theta_{c}, \theta_{a}$}
	
	\SetAlgoLined
	\SetKwProg{Fn}{Function}{}{end}
	\nonl \Fn{Train($\mathcal{V}_x, \beta_{o}, \beta_{c}, \beta_{a}, \gamma_1, \gamma_2$)}{
		
		\For{every epoch}{
			
			\For{every batch of videos $\mathcal{V}_{x}^{B}$}{
				
				$\mathcal{V}_{y}^{B}$ = $O(\mathcal{V}_{x}^{B})$ \mycommfont{$\triangleright$ Acquire private videos}
				
				
				$\theta_{o} \leftarrow \theta_{o} - \beta_{o}\Delta_{\theta_{o}}(L(O) + \gamma_1L(C) - \gamma_2L(A))$
				
				
				$\theta_{c} \leftarrow \theta_{c} - \beta_{c}\Delta_{\theta_{c}}(L(C))$
				
				
				$\theta_{a} \leftarrow \theta_{a} - \beta_{a}\Delta_{\theta_{a}}(L(A))$ 
			}
		}
		
		\KwRet{$\mathbf{X}_{e}$}
	}
	\caption{\small Our Adversarial Training Algorithm.}
	\label{alg:adversarial_training}
\end{algorithm}
\setlength{\floatsep}{0.1cm}

\vspace{-10pt}
\section{Experimental Results}
\vspace{-5pt}

\noindent\textbf{Datasets.} Given the lack of a public dataset containing both human actions and privacy attribute labels on the same videos, we follow the same approach as authors in \cite{wu2020privacy} to train our proposed adversarial framework. Specifically, we perform cross-dataset training using three datasets: the HMDB51 \cite{kuehne2011hmdb}, the VISPR \cite{orekondy2017towards}, and the PA-HMDB51\cite{wu2020privacy}. The VISPR dataset contains 22,167 images annotated with 68 privacy attributes which include: semi-nudity, face, race, gender, skin color, among others. The attributes of a specific image are labeled as ``present" or ``not-present". The HMDB51 dataset comprises 6,849 video clips from 51 action categories, with each category containing at least 101 clips. The Privacy-annotated HMDB51 (PA-HMDB51) is a small subset of the HMDB51 dataset, containing 515 videos, with privacy attribute labels. For each video in PA-HMDB51, there are five attributes annotated on a per-frame basis: skin color, face, gender, nudity, and relationship. Similar to VISPR, the labels are binary and specify if an attribute is present or not in the frame.

\noindent\textbf{Training set.} We train our models using cross-dataset training on HMDB51 and VISPR datasets. Specifically, we exclude the 515 videos in the PA-HMDB51 dataset from HMDB51 and use the remainder videos to train our action recognition component. On the other hand, we use the VISPR dataset with the same five privacy attributes available in the PA-HMDB51 dataset: skin color, face, gender, nudity, and relationship, to train our adversarial component.

\noindent\textbf{Testing set.} We use PA-HMDB51 to test our action recognition and adversarial components. This dataset includes both action and privacy attribute labels.

\noindent\textbf{Training details.} In \Cref{alg:adversarial_training}, we set initial learning rates $\beta_{o}=3\times10^{-3}, \beta_{c}=\beta_{a}=10^{-4}$,  and $\gamma_1=0.7, \gamma_2=0.3$ and applied an exponential learning decay with a decay factor of $0.1$ that is triggered in the epoch $25$. We trained the end-to-end PrivHAR model for 50 epochs, with batch size of 8, and use the Stochastic Gradient Descent (SGD) optimizer to update parameters $\theta_{o}, \theta_{c}, \theta_{a}$. To perform the adversarial attacks during training (adversarial component in \cref{fig:proposedM}), we use the ResNet-50 architecture. Training the PrivHAR for 50 epochs took about 6 hours on 8 Nvidia TITAN RTX GPU with 24 GB of memory.

\vspace{-10pt}
\subsection{Metrics and Evaluation Method}
\label{sec:metrics}
\vspace{-5pt}

To measure the overall performance of PrivHAR, we evaluate the action recognition task and privacy protection separately. First, to test action recognition, we pass the testing videos through our designed camera lens $O(\cdot)$ to obtain the private videos. Next, we use our learned HAR backbone $C(\cdot)$ to get the predicted actions on each private video. Similarly as C3D \cite{tran2015learning}, and RubiksNet \cite{fan2020rubiksnet}, we report the standard average classification accuracy, denoted by $A_C$.

On the other hand, to evaluate privacy protection, we follow the same evaluation protocol adopted by authors in \cite{wu2020privacy}. Specifically, assuming that an attacker has access to the set of private videos acquired with our $O(\cdot)$ and the corresponding privacy attribute labels, then, the attacker can train different CNNs to try to steal sensitive information from the privacy-protected videos acquired with our camera. To empirically verify that our protection is robust to this kind of attack, we separately train ten different classification networks using the private images acquired with our camera, i.e., these CNNs are different from the selected CNN used during training. To train these networks, we use the same training set defined in the previous section and fix our camera component with the optimal learned parameters $\theta_o^{*}$. We use the following architectures: ResNet-$\left\{50,101\right\}$\cite{he2016deep}, Wide-ResNet-$\left\{50,101\right\}$ \cite{zagoruyko2016wide}, MobileNet-V2 \cite{sandler2018mobilenetv2}, Inception-$\left\{\text{V1},\text{V3}\right\}$\cite{szegedy2016rethinking,szegedy2015going}, MNASNet-$\left\{0.5, 0.75, 1.0\right\}$\cite{tan2019mnasnet}. Among these CNNs, eight randomly selected networks start from ImageNet-pretrained weights. The remaining two models were trained from scratch (random initialization) to eliminate the possibility that the initialization with ImageNet weights affects the correct predictions. After training, we evaluate each model on our defined testing set (videos from PA-HMDB51) and select the model with the highest performance. Similar to previous works \cite{dave2022spact,wu2020privacy,wu2018towards,orekondy2017towards}, we adopt the Class-based Mean Average Precision (C-MAP)\cite{orekondy2017towards} to assess the performance of the models. Specifically, we compute the Average Precision (AP) per class, which is the area under the Precision-Recall curve of the privacy-related attribute. Hence, C-MAP corresponds to the average of the AP scores across all the privacy-related attributes. We also denote C-MAP as $A_A$ in our experiments (lower is better).

To measure image degradation, we use the structural similarity index (SSIM) metric \cite{hore2010image}. Large values of SSIM indicate high quality. Thus, in general, we expect to achieve the minimum SSIM values while achieving high $A_C$ and low $A_A$ for HAR and Adversarial accuracy, respectively. Besides, we combine the two accuracy metrics ($A_C$ and $A_A$) into one using the harmonic mean as:
{\small
	\setlength{\belowdisplayskip}{4pt}
	\setlength{\abovedisplayskip}{4pt}
	\begin{equation}
		P = \frac{2}{\frac{1}{A_C}+\frac{1}{1-A_A}}=\frac{2A_C (1-A_A)}{1-A_A+A_C},
	\end{equation}
}and we expect to achieve the maximum $P$ value.

\subsection{Simulation Experiments}

\setlength{\tabcolsep}{4pt}
\begin{table*}[t]
	\begin{subtable}[h]{0.45\linewidth}
		\centering
		\caption{\scriptsize Ablation Study}
		\resizebox{\columnwidth}{!}{%
			\setlength{\tabcolsep}{3pt}
			\def\arraystretch{0.8}%
			\begin{tabular}{@{}lcccc@{}}
				\multicolumn{5}{c}{C3D Backbone}                                                                                     \\ \midrule
				Experiment     & SSIM$\downarrow$ & $A_C\uparrow$                  & $A_A\downarrow$                & $P\uparrow$                   \\ \midrule
				No-Adversarial & 0.603            & 51.1                           & 69.1                           & 38.6                            \\
				No-TSM         & 0.612            & 59.9                           & 69.7                           & 40.2                            \\ \midrule
				Zernike-50     & 0.643            & 58.3                           & 70.5                           & 39.2                            \\
				Zernike-100    & 0.629            & 58.8                           & 69.3                           & 40.4                            \\
				Zernike-200    & 0.612            & \textbf{63.3} & \textbf{68.9} & \textbf{41.52} \\ \midrule
				\multicolumn{5}{c}{RubiksNet Backbone}                                                                               \\ \midrule
				No-Adversarial & 0.592            & 57.6                           & 68.2                           & 40.9                            \\
				No-TSM         & 0.599            & 72.3                           & 67.6                           & 44.6                            \\ \midrule
				Zernike-50     & 0.618            & 70.2                           & 69.2                           & 42.8                            \\
				Zernike-100    & 0.601            & 71.9                           & 68.4                           & 43.9                            \\
				Zernike-200    & 0.588            & \textbf{73.8} & \textbf{66.5} & \textbf{46.1}  \\ \bottomrule
		\end{tabular}}
		\label{tab:ablation_C3D}
	\end{subtable}
	\hfill
	\begin{subtable}[h]{0.51\linewidth}
		\centering
		\caption{\scriptsize Comparisons.}
		\resizebox{\columnwidth}{!}{%
			\setlength{\tabcolsep}{3pt}
			\def\arraystretch{1.047}%
			\begin{tabular}{@{}lcccc@{}}
				\toprule
				Methods                                                           & SSIM$\downarrow$ & $A_C\uparrow$ & $A_A\downarrow$ & $P\uparrow$         \\ \midrule
				No-privacy (C3D)                                                         & 1.0       & 71.1  & 76.1  & 35.8         \\
				No-privacy (RubiksNet)                                                         & 1.0       & 85.2  & 76.1  & 37.3         \\ \midrule
				Low-resolution \cite{ryoo2018extreme}                                                     & 0.686    & 48.3  & 70.9  & 36.3                \\
				Lens in \cite{Hinojosa_2021_ICCV}-RubiksNet  & 0.608    & 52.4  & 69.4  & 38.6                \\
				Defocus \cite{pittaluga2015privacy}                                                            & 0.688    & 62.1  & 72.5  & 38.1              \\ \midrule
				PDAR-GRL \cite{wu2020privacy}   & -         & 63.3  & 70.5  & 40.2          \\
				PDAR-K-Beam  \cite{wu2020privacy}          & -         & 63.5  & 69.3  & 41.4          \\
				PDAR-Entropy  \cite{wu2020privacy}         & -         & {\ul67.3}  & 70.3  & 41.2          \\ \midrule
				\textbf{PrivHAR-C3D}                                            & 0.612    & 63.3  & {\ul 68.9}  & {\ul 41.7} \\
				\textbf{PrivHAR-RubiksNet}                                      & 0.588    & \textbf{73.8}  & \textbf{66.5}  & \textbf{46.1} \\ \bottomrule
		\end{tabular}}
		\label{tab:comparison}
	\end{subtable}
	\vspace{5pt}
	\caption{\small Quantitative Results. (a) Multiple ablation studies of our method for two different HAR backbones, C3D and RubiksNet: each component in \cref{fig:proposedM} is trained separately (No-Adversarial); not using the TSM matrices to preserve temporal information (No-TSM); 50, 100, and 200 Zernike polynomials to design our lens. (b) Comparison of our method (\textbf{PrivHAR}) with: three additional privacy-preserving approaches: defocusing, low-resolution cameras, and the lens used in \cite{Hinojosa_2021_ICCV}; and the privacy-preserving deep action recognition (PDAR) framework with different learning approaches (GRL, K-Beam, and Entropy) \cite{wu2020privacy}. Accuracy values are reported in percentage.}
	\label{tab:all_quantitative_results}
\end{table*}
\setlength{\tabcolsep}{1.4pt}

\noindent\textbf{Ablation Studies.} We conduct multiple experiments to investigate different configurations for our adversarial approach. We show the quantitative results of our ablations studies in Table \ref{tab:all_quantitative_results} (a), for C3D and RubiksNet. We first train the optical and action recognition components to obtain privacy-preserving videos and perform HAR on them. Then, we fix the optical component and train the adversarial CNN to recover the privacy attributes from the videos. We refer to this experiment as `No-adversarial' in the Table \ref{tab:all_quantitative_results} (a). Note that this approach is similar to the prior work \cite{Hinojosa_2021_ICCV} but on a different vision task. In our second experiment (No-TSM), we test the performance of our proposed PrivHAR with $q=200$ Zernike coefficients when not using TSMs to preserve the temporal information. We can observe from the table that, in general, the $A_C$ decreases, which evidences the importance of using TSMs to preserve temporal information. The third experiment consists of training our adversarial framework with a different number of Zernike coefficients. Specifically, we trained our PrivHAR using $q=50$, $q=100$, and $q=200$ Zernike coefficients, see \cref{eq:zernike_model}. In general, increasing the number of Zernike coefficients leads to better encoding; hence the $A_C$ value increases while the SSIM decreases. However, memory consumption also increases since we need to store all the Zernike bases. In general, we use $q=200$ Zernike coefficients as a default value in all other experiments. The tables show that the best HAR backbone for our proposed PrivHAR network is RubiksNet. We observed that when using RubiksNet, PrivHAR achieves higher distortions (lower SSIM) affecting the performance of the adversarial component while achieving high action recognition accuracy. We empirically verify that RubiksNet is better at preserving the temporal information than C3D; hence it performs better even with high image distortions. Besides, we observed that TSM helps more the C3D backbone, which is more affected by the distortions generated by our lens. 

\noindent\textbf{Attribute Estimator Network Performance.} The values of $A_A$ reported in the tables corresponds to the C-MAP obtained by the model with highest performance on our testing set, as described in Section \ref{sec:metrics}. To analyze the performance of the attribute estimator networks, and hence our privacy protection, we plot the receiver operating characteristic (ROC) and Precision-Recall (PR) curves. In our supplementary document, we show the ROC and PR curves of the attribute estimator network which achieves the best performance on the privacy-preserving images/videos acquired with our camera. Specifically, considering the area under curve (AUC) of the PR curves, we obtain an average precision (AP) of 0.94, 0.72, 0.97, 0.52, 0.18 for skin color, face, gender, nudity, and relationship, respectively. These values of AP are very close to those obtained by a random classifier (null hypothesis), which are 0.95, 0.71, 0.97, 0.58, 0.17. Therefore, based on the Fisher's exact test \cite{upton1992fisher}, the best attribute estimator network on our privacy-protected images is not significantly different from the random classifier ($p$-value$<0.01$).

\begin{figure}[t]
	\centering
	\includegraphics[width=0.95\columnwidth]{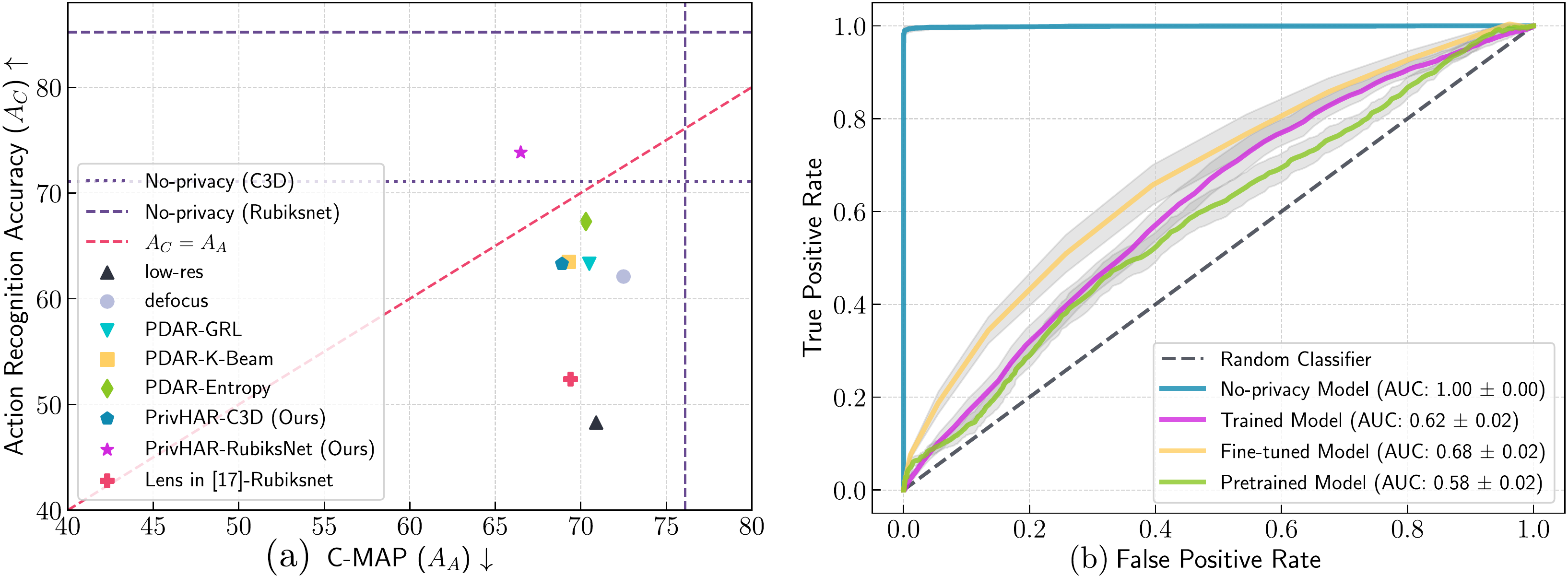}\vspace{-3pt}
	\caption{\small (a) Trade-off between privacy protection and action recognition on PA-HMDB51. Vertical and horizontal, dashed and dotted, purple lines indicate $A_A$ and $A_C$ on the original non-privacy videos, using RubiksNet and C3D backbones for HAR, respectively. The red dashed line indicates where $A_A=A_C$. (b) Face recognition performance on private images (from LFW \cite{Huang2012a} dataset) acquired with our optimized lens.}\vspace{5pt}
	\label{fig:scatter_plot}
\end{figure}

\noindent\textbf{Comparison with other methods}. We compare our proposed PrivHAR with two traditional privacy-preserving approaches: low-resolution \cite{ryoo2017privacy} and defocusing cameras \cite{pittaluga2015privacy}. We simulate both types of cameras and perform a similar training as shown in \cref{fig:proposedM}. To implement the low-resolution approach, we manually downsampled the images with a resolution of $16 \times 16$. In addition, we compare our proposed PrivHAR with the privacy-preserving deep action recognition (PDAR) framework with different learning approaches (GRL, K-Beam, and Entropy) \cite{wu2020privacy}. We present the quantitative results in Table \ref{tab:all_quantitative_results} (b), where all methods use the C3D backbone for HAR if not otherwise specified. We also include our PrivHAR using RubiksNet for comparison. Furthermore, we use the lens designed in \cite{Hinojosa_2021_ICCV}, which was optimized for human pose estimation and did not consider adversarial attacks during training, for distort the videos and then perform HAR on them. This approach obtains an $A_C=52.4\%$ using RubiksNet, which is $21.4\%$ lower than our PrivHAR-RubiksNet results. In addition, the trade-off between privacy protection and action recognition is visualized in Fig. \ref{fig:scatter_plot} (a), which shows PrivHAR obtains the best privacy while maintaining high accuracy.

\noindent\textbf{Face recognition performance}. We follow the same face recognition validation on private images acquired by the optimized lens as the prior work in \cite{Hinojosa_2021_ICCV}. Specifically, we use an implementation of the face recognition network ArcFace \cite{deng2019arcface}, train on Microsoft Celeb (MS-Celeb-1M) \cite{guo2016ms} and test on LFW \cite{Huang2012a} datasets. \Cref{fig:scatter_plot} (b) show the ROC curves for each testing approach: ``No-privacy Model'' uses the pretrained ArcFace model on the original (non-private) images; ``Pretrained model'' uses the pretrained ArcFace model on the private version of each dataset; ``Trained model'' uses an ArcFace model trained from scratch using the private version of the MS-Celeb-1M dataset; ``Fine-tuned Model'' uses a pretrained ArcFace model fine-tuned with the private version of the MS-Celeb-1M dataset. From the figure, we can conclude that the ArcFace model does not perform well on the images generated by our designed lens as the best performance is achieved by the fine-tuned model (AUC$=0.68$), which is still close to random classifier's performance. See results with others datasets in our supplementary.

\begin{figure*}[t]
	\centering
	\includegraphics[width=0.96\linewidth]{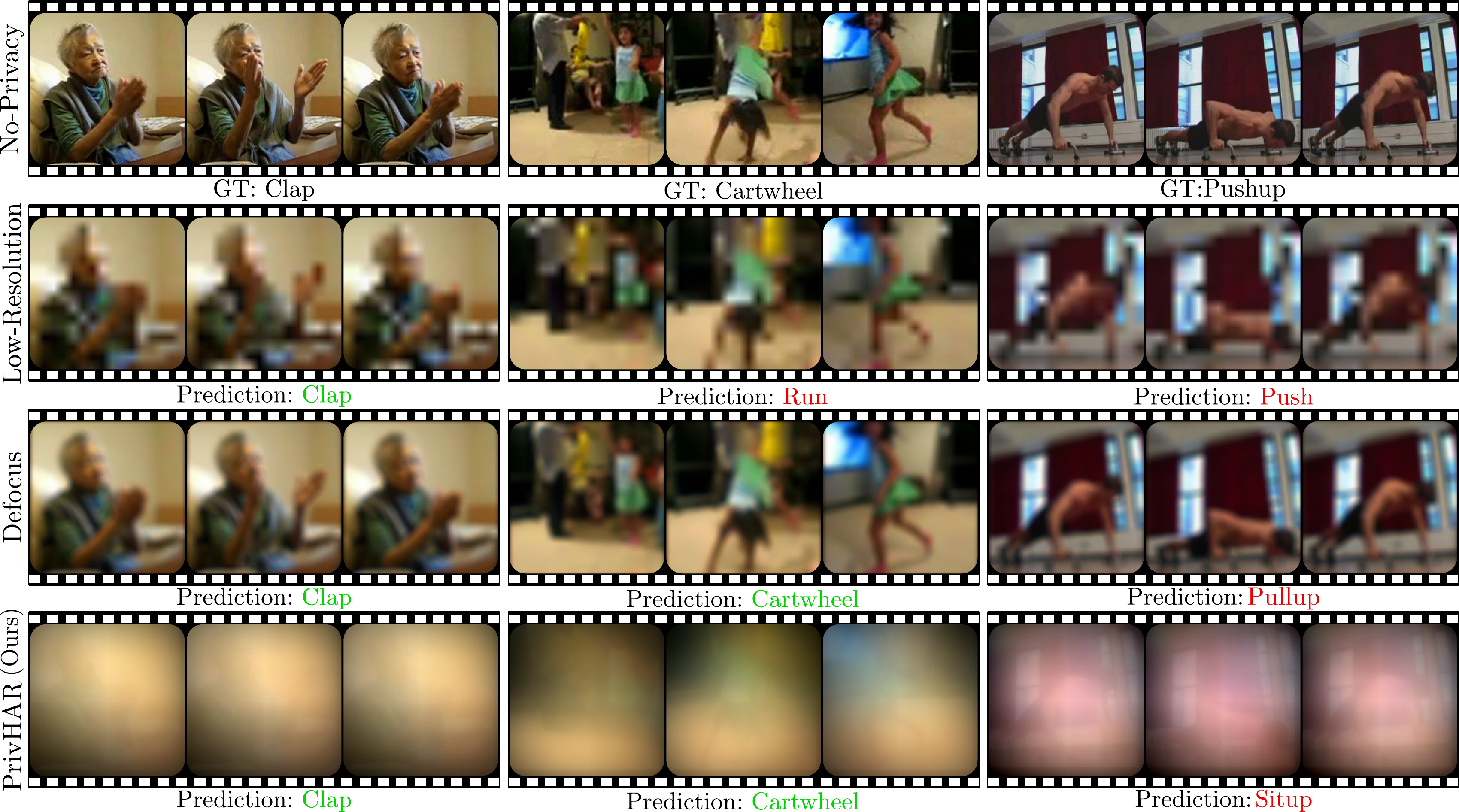}\vspace{-8pt}
	\caption{\small Qualitative Results on PA-HMDB51. Each row shows standard no-privacy videos and ground truth (GT) labels (top); and predictions from our optimized lens (PrivHAR-RubiksNet, bottom) to low-resolution (second) and defocus (third) cameras.}\vspace{7pt}
	\label{fig:visual_results}
\end{figure*}

\noindent\textbf{Qualitative Results}. We qualitatively compare our approach with low resolution and defocusing cameras in \cref{fig:visual_results}. We show results on three example videos from the PA-HMDB51 dataset. The first row of the figure shows the non-privacy video acquired using a standard lens and the ground truth (GT) of the actions for reference. As observed, our lens achieves a higher distortion but still performs action recognition. The last video shows a failure case of our method.

\begin{figure}[t]
	\centering
	\includegraphics[width=0.96\columnwidth]{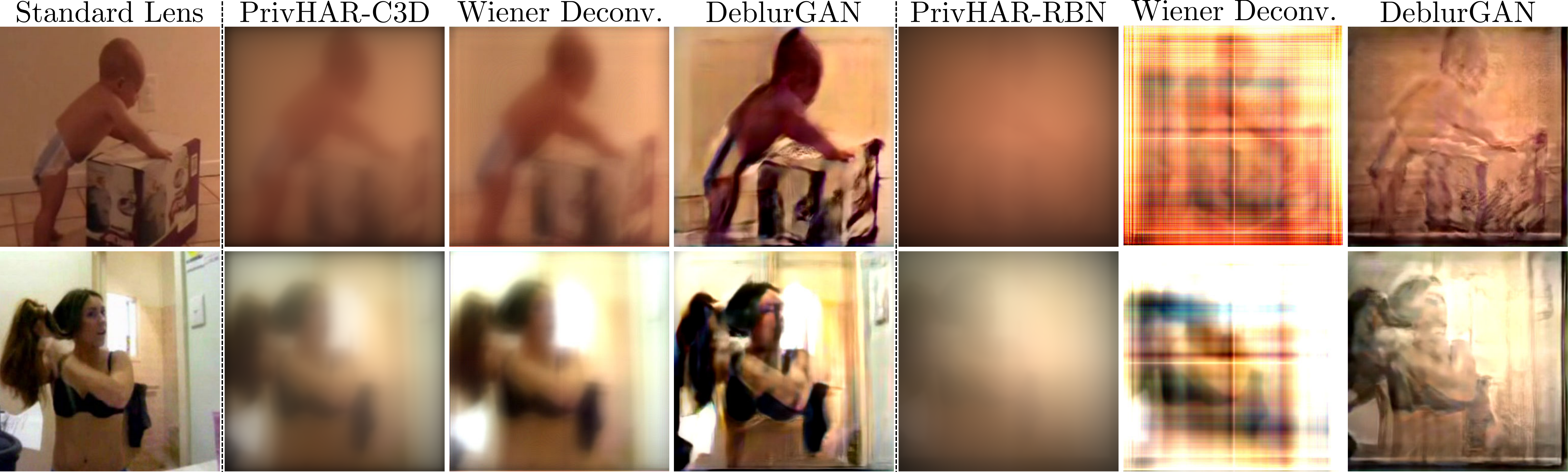} \vspace{-8pt}
	\caption{\small Deconvolution of private images acquired with our optimized lens using C3D and RubiksNet (RBN) backbones in PrivHAR. The images acquired with our lens are robust to deconvolution, and DeblurGAN cannot recover people's identities.}\vspace{7pt}
	\label{fig:blind_deconv}
\end{figure}

\noindent\textbf{Deconvolution Attack.} Suppose the attacker has access to the camera or a large collection of acquired images with our proposed camera. In that case, the attacker could use deconvolution methods (blind and non-blind) on our distorted images to recover people's identities. To test the robustness of our designed lens to deconvolution attacks, we assume both scenarios: having access to the camera, we can easily get the PSF (by imaging a point of source light) and hence use a non-blind deconvolution method, e.g. the Wiener deconvolution; on the other hand, not having access to the camera but a large collection of our distorted images then we can train a blind deconvolution network, e.g. DeblurGAN \cite{kupyn2019deblurgan}. We describe the training details in our supplementary document. In \cref{fig:blind_deconv} we show the results with two video frames from the HMDB51 dataset with people near the camera. We observed that the distortion achieved by PrivHAR-RubiksNet (RBN) is significantly higher than C3D; hence it is more difficult for DeblurGAN and Wiener deconvolution to recover the scene. In both cases, using C3D or RBN, the distortion is sufficient to avoid recovering face details, and the people's identity is protected. However, some attributes are visible in the recovered scene when using PrivHAR-C3D. It is possible to obtain a lens with C3D that provides more distortion; however, the HAR accuracy could be affected.

\begin{figure}[t]
	\centering
	\includegraphics[width=1\linewidth]{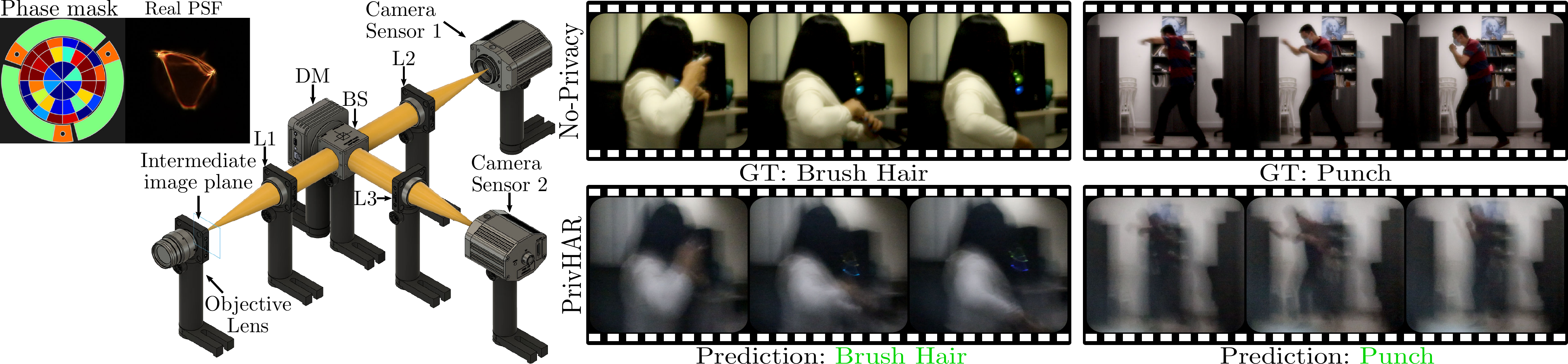} \vspace{-15pt}
	\caption{\small Experimental setup scheme and some results on acquired videos. The deformable mirror configuration and characterized PSF are shown in the upper left corner. The right column shows the non-privacy and private videos obtained with our camera.
	}\vspace{5pt}
	\label{fig:Real}
\end{figure}

\vspace{-5pt}
\subsection{Hardware Experiments}

To demonstrate the PrivHAR's capability of action recognition, we conduct experimental validations acquiring four human actions: jump, clap, punch, and hair brush in our Lab. We emulate the lens designed with our PrivHAR adversarial framework using a deformable mirror-based 4f system \cite{marquez2019compressive,marquez2021snapshot}. We first train our system using $q=15$ Zernike coefficients and then load the learned coefficients to the deformable mirror an calibrate the PSF. After calibration we obtained the following learned Zernike coefficients: {\small $\{\alpha_1=\alpha_2=\alpha_3=0, \alpha_4=-0.45, \alpha_5=0.36, \alpha_6=0.24, \alpha_7=0.6, \alpha_8=-0.4, \alpha_9=-0.11, \alpha_{10}=0.69, \alpha_{11}=-0.31, \alpha_{12}=-0.15, \alpha_{13}=-0.70, \alpha_{14}=-0.85, \alpha_{15}=0.38\}$}. The resulting PSF and the used phase mask are presented in Fig. \ref{fig:Real}(Left). Finally, we placed our proof-of-concept system on a movable table to take it out of our Lab and acquire real outdoor images. In Fig. \ref{fig:Real}(Right), we show the human action recognition for two video sequences recorded by our 4F-based system. The ground truth and the private version were illustrated in the first and second rows, respectively. Outdoor system configuration, additional qualitative and quantitative results, and detailed description of the proof-of-concept system can be found in the supplement.






\vspace{-10pt}
\section{Discussion and Conclusion}
\vspace{-5pt}
We present PrivHAR, a framework for detecting human actions from a privacy-preserving lens. Our framework consists of three components: the hardware component that comprises a camera with a privacy-preserving lens, whose parameters are learned during training and its main function is to obscure sensitive private information; the action recognition component that aims to preserve temporal information using temporal similarity matrices and performs HAR on the degraded video; and the adversarial component, which performs five attacks to the private videos seeking to recover the hidden attributes.


\noindent\textbf{Limitations.} One limitation of our simulated experiments is that we test our approach on a relatively small set due to the lack of a public dataset containing human actions and privacy attribute labels on the same videos. As future work, we plan to build a video dataset using our proposed optical system, which allows us to acquire both RGB and private videos. In addition,  the deformable mirror is the main limitation of the proof-of-concept optical system. This device can only use $q = 15$ Zernike Polynomials, limiting the scene's level of distortion. For now, our small-scale tests show results consistent with our extensive experiments.

\noindent\textbf{Conclusion.} We extensively evaluated and experimentally validated our approach in simulations and a hardware prototype. Our qualitative and quantitative results indicate a trade-off between image degradation and HAR accuracy. Our optics modeling can generally be integrated into an embedded chipset or used as a software-only solution by applying the image degradation post-acquisition to deploy a less secure system. However, we show that the learned lens can be deployed as a camera, which provides a higher security layer. One could connect it to an Nvidia Jetson for real-time privacy-preserving HAR. 






%
%
\bibliographystyle{splncs04}
\bibliography{egbib}
\end{document}